\documentclass{INTERSPEECH2023}
\usepackage{xcolor}
\usepackage{multirow}
\usepackage{pdflscape} 
\usepackage{xspace}
\usepackage{adjustbox}
\usepackage{rotating}
\usepackage{soul} 

\newcommand{\muavic}{MuAViC}
\newcommand{\hs}[1]{\hspace{#1\tabcolsep}}

\interspeechcameraready 

\title{\muavic: A Multilingual Audio-Visual Corpus for Robust Speech Recognition and Robust Speech-to-Text Translation}
\name{Mohamed Anwar, Bowen Shi, Vedanuj Goswami, Wei-Ning Hsu, Juan Pino, Changhan Wang}
\address{Meta AI}
\email{\{anwarvic,bshi,vedanuj,wnhsu,juancarabina,changhan\}@meta.com}

\begin{document}

\maketitle
\begin{abstract}
We introduce \muavic, a multilingual audio-visual corpus for robust speech recognition and robust speech-to-text translation providing 1200 hours of audio-visual speech in 9 languages. It is fully transcribed and covers 6 English-to-X translation as well as 6 X-to-English translation directions. To the best of our knowledge, this is the first open benchmark for audio-visual speech-to-text translation and the largest open benchmark for multilingual audio-visual speech recognition. Our baseline results show that \muavic~is effective for building noise-robust speech recognition and translation models. We make the corpus available at \url{https://github.com/facebookresearch/muavic}.
\end{abstract}

\section{Introduction}

Visual perception of lip movement plays a crucial role in supplementing audio signals for a better understanding of spoken languages~\cite{Sumby1954VisualCT}. Its immunity to acoustic noise makes it a valuable asset in building noise-robust speech processing systems. Audio-visual speech recognition (AVSR), which transcribes spoken utterances using both audio and visual inputs, has been shown to be effective in improving the robustness of speech recognition~\cite{Afouras2018DeepAS,shi22_interspeech}. The application of deep learning has greatly improved the performance of AVSR systems, as evidenced by the state-of-the-art (SOTA) models in the widely used LRS3-TED public benchmark. Typically, SOTA systems~\cite{shi22_interspeech} are able to lower word error rate (WER) by 3 times
compared to its audio-only counterpart under challenging conditions with 0~dB babble noise.

In spite of the encouraging progress mentioned earlier, modern AVSR models are heavily benchmarked under a monolingual setup, with the majority being based on English. Non-English audio-visual datasets are orders of magnitude smaller, with a common dataset, CMU-MOSEAS~\cite{bagher-zadeh-etal-2020-cmu} providing less than 20 hours of audio-visual speech for each of the 4 non-English languages it covers, which is at least 20 times smaller than the English-based LRS3-TED corpus. The lack of large-scale audio-visual datasets also impedes the development of audio-visual speech-to-text translation (AVST), which is only simulated using synthetic data 
in prior works, but is expected to bring a noise-robust effect similar to AVSR, as will be demonstrated in this paper.
With the increasing popularity of self-supervised audio-visual pre-training approaches~\cite{shi2021learning}, proper benchmarking of multilingual AVSR and AVST models will enable multilingual pre-training, which allows more effective use of large-scale multilingual audio-visual data such as AV-Speech~\cite{Ephrat2018LookingTL}.

\begin{table}[t]
    \centering
    \fontsize{7.5}{10}\selectfont
    \caption{Duration and speaker statistics for \muavic. $^\dagger$ Train set example is either human-labeled (``H") or pseudo-labeled (``P") by a machine translation model, M2M-100 \textit{418M}~\cite{fan2021beyond}. 
    }
    \begin{tabular}{lrrrrrr}
    \toprule
    & \multicolumn{3}{c}{Hours} & \multicolumn{3}{c}{Speakers} \\
    \cmidrule(lr){2-4} \cmidrule(lr){5-7}
    & Train (H+P)$^\dagger$ & Dev & Test & Train & Dev & Test \\
    \midrule[\heavyrulewidth]
    \multicolumn{7}{c}{\it Audio-Visual Speech Recognition} \\
    En~\cite{afouras2018lrs3,shi22_interspeech} & 436 + 0 & 1.1 & 0.8 & 4.7K & 0.9K & 0.4K \\
    Ar                                          &  16 + 0 & 1.5 & 1.2 & 95   & 7    & 7    \\
    De                                          &  10 + 0 & 1.6 & 1.5 & 53   & 9    & 9    \\
    El                                          &  25 + 0 & 2.3 & 2.0 & 113  & 10   & 8    \\
    Es                                          & 178 + 0 & 1.6 & 1.7 & 987  & 16   & 12   \\
    Fr                                          & 176 + 0 & 1.8 & 1.5 & 948  & 12   & 10   \\
    It                                          & 101 + 0 & 1.9 & 1.9 & 487  & 8    & 8    \\
    Pt                                          & 153 + 0 & 1.5 & 1.8 & 810  & 9    & 13   \\
    Ru                                          &  49 + 0 & 1.7 & 1.8 & 238  & 7    & 9    \\
    \midrule[\heavyrulewidth]
    \multicolumn{7}{c}{\it En-X Audio-Visual Speech-to-Text Translation} \\
    En-El & 17 + 420 & 0.2 & 0.5 & \multirow{6}{*}{4.7K} & \multirow{6}{*}{0.9K} & \multirow{6}{*}{0.4K} \\
    En-Es & 21 + 416 & 0.2 & 0.6 &                                                                       \\
    En-Fr & 21 + 416 & 0.2 & 0.6 &                                                                       \\
    En-It & 20 + 417 & 0.2 & 0.6 &                                                                       \\
    En-Pt & 18 + 419 & 0.2 & 0.5 &                                                                       \\
    En-Ru & 20 + 417 & 0.2 & 0.6 &                                                                       \\
    \midrule[\heavyrulewidth]
    \multicolumn{7}{c}{\it X-En Audio-Visual Speech-to-Text Translation} \\
    El-En & 8 + {} {} 17  & 2.3 & 2.0 & 113 & 10 & 8  \\
    Es-En & 64 + 114      & 1.6 & 1.7 & 987 & 16 & 12 \\
    Fr-En & 45 + 131      & 1.8 & 1.6 & 948 & 12 & 10 \\
    It-En & 48 + {} {} 53 & 1.9 & 1.9 & 487 & 8  & 8  \\
    Pt-En & 53 + 100      & 1.5 & 1.8 & 810 & 9  & 13 \\
    Ru-En & 8 + {} {} 41  & 1.7 & 1.8 & 238 & 7  & 9  \\
\bottomrule
\end{tabular}
\label{tab:stats}
\end{table}

\begin{table*}[t]
    \centering
    \fontsize{7.5}{10}\selectfont
    \caption{Results for English speech recognition and En-X speech translation in a clean environment (A: audio, AV: audio+video). 
    }
    \begin{tabular}{llcrrrrrrrr}
    \toprule
    \multirow{2}{*}{ID} & \multirow{2}{*}{Model} & \multirow{2}{*}{Mode} & \multicolumn{7}{c}{Target} & \multirow{2}{*}{Avg} \\
    \cmidrule(lr){4-10}
     & & & En & El & Es & Fr & It & Pt & Ru & \\
    \midrule[\heavyrulewidth]
    \multicolumn{11}{c}{\it Speech Recognition, Test WER $\downarrow$} \\
    \multirow{2}{*}{\texttt{A1}} & \multirow{2}{*}{Monolingual AV-HuBERT \cite{shi22_interspeech}} & A  & 2.5  & - & - & - & - & - & - & - \\
    &                                                                                              & AV & 2.3  & - & - & - & - & - & - & - \\
    \midrule[\heavyrulewidth]
    \multicolumn{11}{c}{\it Text-To-Text Translation, Test BLEU $\uparrow$} \\
    \texttt{A2} & Bilingual Transformer \textit{base}                                          & - & - & 25.8 & 29.5 & 27.0 & 22.6 & 23.9 & 17.2 & 24.3 \\
    \texttt{A3} & M2M-100 \textit{418M}, multilingual, unconstrained data~\cite{fan2021beyond} & - & - & 24.5 & 28.7 & 25.6 & 21.8 & 22.2 & 15.8 & 23.1 \\
    \midrule[\heavyrulewidth]
    \multicolumn{11}{c}{\it Speech-To-Text Translation, Test BLEU $\uparrow$} \\
     & \multirow{2}{*}{\texttt{A1}+\texttt{A2}} & A  & - & 25.1 & 28.9 & 26.7 & 21.9 & 23.8 & 16.9 & 23.9 \\
     &                                          & AV & - & 25.5 & 29.1 & 26.5 & 22.3 & 23.6 & 16.8 & 24.0 \\
    \cmidrule(lr){1-11}
      & \multirow{2}{*}{\texttt{A1}+\texttt{A3}} & A  & - & 23.8 & 27.8 & 25.4 & 21.3 & 22.0 & 15.2 & 22.6 \\
      &                                          & AV & - & 24.1 & 28.2 & 25.2 & 21.4 & 22.0 & 15.3 & 22.7 \\
    \cmidrule(lr){1-11}
    & \multirow{2}{*}{Bilingual AV-HuBERT} & A  & - & 23.0 & 27.5 & 25.1 & 20.7 & 20.1 & 14.7 & 21.9 \\
    &                                      & AV & - & 23.4 & 26.6 & 25.3 & 20.7 & 20.5 & 14.6 & 21.9 \\

    \bottomrule
    \end{tabular}
    \label{tab:en_x}
\end{table*}

In this paper, we present \muavic, a multilingual audio-visual corpus for robust speech recognition and robust speech-to-text translation.
This corpus is sourced from TED and TEDx talks including total 1200 hours of transcribed audio-visual speech from over 8000 speakers in 9 languages: English (En), Arabic (Ar), German (De), Greek (El), Spanish (Es), French (Fr), Italian (It), Portuguese (Pt) and Russian (Ru). This makes \muavic~the largest open benchmark so far for multilingual audio-visual speech recognition (AVSR) and lipreading. In addition, we provide text translations and establish baselines for 6 English-to-X translation as well as 6 X-to-English translation directions. To the best of our knowledge, \muavic~is the first publicly available corpus for audio-visual speech-to-text translation.


\section{Related Works}


\textbf{Audio-visual speech recognition}
Audio-visual speech recognition (AVSR) has been a popular research topic for the speech community for a long time. Most of the early AVSR datasets, such as GRID~\cite{Cooke2006AnAC}, were collected in a controlled environment where subjects were required to recite pre-defined phrases in a laboratory-like setting (for example, with uniform illumination conditions). These datasets also have limitations in terms of visual perspectives and often consist of small-scale data with limited vocabulary size and number of speakers. The first large-scale ``in-the-wild'' AVSR dataset was LRW~\cite{Chung2016LipRI}, which contains 173 hours of speaking video clips of 500 isolated words from BBC broadcasts. In recent years, there has been a growing trend towards collecting sentence-level audio-visual speech data, such as LRS3-TED~\cite{afouras2018lrs3,Afouras2018DeepAS}, which is more natural and widely accessible. LRS3-TED is the largest publicly available AVSR dataset and contains approximately 437 hours of English speaking videos from over 5000 speakers.

The majority of the widely-used AVSR datasets are in English, while non-English AVSR datasets~\cite{bagher-zadeh-etal-2020-cmu,Zhao2019ACS,ivanko-etal-2022-rusavic} are significantly smaller in size. For example, the largest AVSR dataset in Mandarin Chinese, CMLR~\cite{Zhao2019ACS}, comprises only 86 hours of utterances from 11 speakers. The current largest multilingual AVSR dataset, CMU-MOSEAS~\cite{bagher-zadeh-etal-2020-cmu}, contains total 68 hours of speech in Spanish, Portuguese, German, and French. Furthermore, it lacks cross-lingual translations, hindering its use for audio-visual speech-to-text translation (AVST).

It is noteworthy that there exist large-scale, unlabeled multilingual audio-visual speech datasets, such as VoxCeleb2~\cite{Chung2018VoxCeleb2DS} (2442 hours) and AV-Speech~\cite{Ephrat2018LookingTL} (around 4700 hours). They have been widely utilized for various non-transcription tasks, such as speaker verification~\cite{Chung2018VoxCeleb2DS,Chung2020InDO} and source separation~\cite{Ephrat2018LookingTL}, or as a source of pre-training data for self-supervised audio-visual representation learning~\cite{shi2021learning,hsu2022uhubert,Afouras2019ASRIA}. However, they are not directly applicable to AVSR and AVST due to the lack of transcriptions.

\noindent\textbf{Speech-to-text translation} Speech-to-text translation (ST) or spoken language translation (SLT) has become an increasingly popular topic of speech research with the recent emergence of open ST benchmarks. Almost all the ST corpora were created by adding translations of transcriptions to pre-existing speech recognition corpora, where speech data was collected from various sources, such as phone call recordings~\cite{post-etal-2013-improved}, read speech recordings~\cite{kocabiyikoglu-etal-2018-augmenting,wang-etal-2020-covost,wang21s_interspeech} and public speech recordings~\cite{di2019must,iranzo2020europarl,cattoni2021must,salesky2021mtedx}.
Despite the occasional availability of video tracks in the original data sources of these corpora, their creators have focused on only the audio modality and included only audio data in the corpora. 
Due to the lack of labeled audio-visual corpora, audio-visual speech-to-text translation remains a largely uncharted research area.

\begin{table*}[t]
    \centering
    \fontsize{7.5}{10}\selectfont
    \caption{Results for non-English speech recognition and X-En speech-to-text translation in a clean environment (A: audio, AV: audio+video)
    }
    \begin{tabular}{llcrrrrrrrrrr}
    \toprule
    \multirow{2}{*}{ID} & \multirow{2}{*}{Model} & \multirow{2}{*}{Mode} & \multicolumn{8}{c}{Source} & \multirow{2}{*}{Avg} \\
    \cmidrule(lr){4-11}
     & & & Ar & De & El & Es & Fr & It & Pt & Ru & \\
    \midrule[\heavyrulewidth]
    \multicolumn{12}{c}{\it Speech Recognition, Test WER $\downarrow$} \\
    & Transformer, monolingual \cite{salesky2021mtedx}                        & A & 104.4 & 111.1 & 109.5 & 46.4 & 45.6 & 48.0 & 54.8 & 74.7 & 74.3 \\
    & Hybrid LF-MMI, monolingual \cite{salesky2021mtedx}                      & A & 80.8  & 42.3  & 25.0  & 16.2 & 19.4 & 16.4 & 20.2 & 28.4 & 31.1 \\
    & Whisper, multilingual, unconstrained data \cite{radford2022robust}      & A & 91.5  & 24.8  & 25.4  & 12.0 & 12.7 & 13.0 & 15.5 & 31.1 & 28.2 \\
    \cmidrule(lr){1-12}
    \multirow{2}{*}{\texttt{B1}} & \multirow{2}{*}{Monolingual AV-HuBERT}  & A  & 99.3 & 61.1 & 35.1 & 16.5 & 24.4 & 19.3 & 23.0 & 33.3 & 39.0 \\
                                 &                                         & AV & 98.5 & 52.4 & 26.3 & 15.9 & 23.7 & 18.5 & 19.4 & 30.0 & 35.6 \\
    \cmidrule(lr){1-12}
    \multirow{2}{*}{\texttt{B2}} & \multirow{2}{*}{Multilingual AV-HuBERT} & A  & 67.7 & 46.5 & 40.4 & 30.6 & 27.0 & 19.3 & 19.8 & 37.6 & 36.1 \\
                                 &                                         & AV & 69.3 & 47.2 & 41.2 & 16.2 & 19.0 & 19.8 & 19.9 & 38.0 & 33.8 \\
    \midrule[\heavyrulewidth]
    \multicolumn{12}{c}{\it Text-To-Text Translation, Test BLEU $\uparrow$} \\
    \texttt{B3} & Bilingual Transformer \textit{base}                                          & - & - & - & 13.9 & 28.2 & 33.8 & 27.2 & 31.9 & 14.2 & 27.0 \\
    \texttt{B4} & M2M-100 \textit{418M}, multilingual, unconstrained data~\cite{fan2021beyond} & - & - & - & 25.9 & 29.5 & 34.9 & 30.7 & 33.6 & 19.1 & 29.0 \\
    \midrule[\heavyrulewidth]
    \multicolumn{12}{c}{\it Speech-To-Text Translation, Test BLEU $\uparrow$} \\
    & Whisper, multilingual, unconstrained data \cite{radford2022robust} & A & - & - & 24.2 & 28.9 & 34.5 & 29.2 & 32.6 & 16.1 & 29.9 \\
    \cmidrule(lr){1-12}
    & \multirow{2}{1.4cm}{\texttt{B1}+\texttt{B3}} & A  & \multirow{2}{*}{-} & \multirow{2}{*}{-} & 10.7 & 21.2 & 24.2 & 20.3 & 23.5 & 10.7 & 18.4 \\
    &                                              & AV &                    &                    & 11.3 & 21.1 & 24.3 & 20.8 & 24.9 & 11.0 & 18.9 \\
    \cmidrule(lr){1-12}
    & \multirow{2}{1.4cm}{\texttt{B2}+\texttt{B3}} & A  & \multirow{2}{*}{-} & \multirow{2}{*}{-} & 7.6 & 9.2  & 12.8 & 19.3 & 23.1 & 8.7 & 13.4 \\
    &                                              & AV &                    &                    & 7.5 & 20.2 & 24.4 & 18.6 & 23.2 & 8.6 & 17.1 \\
    \cmidrule(lr){1-12}
    & \multirow{2}{1.4cm}{\texttt{B1}+\texttt{B4}} & A  & \multirow{2}{*}{-} & \multirow{2}{*}{-} & 15.9 & 23.1 & 25.7 & 23.4 & 24.2 & 13.6 & 21.0 \\
    &                                              & AV &                    &                    & 18.4 & 23.0 & 26.1 & 23.8 & 26.3 & 13.9 & 21.9 \\
    \cmidrule(lr){1-12}
    & \multirow{2}{*}{\centering Bilingual AV-HuBERT} & A  & \multirow{2}{*}{-} & \multirow{2}{*}{-} & 7.3 & 20.4 & 26.4 & 19.5 & 24.2 & 9.2 & 17.8 \\
    &                                                 & AV &                    &                    & 7.2 & 20.2 & 25.0 & 19.4 & 24.1 & 9.0 & 17.5 \\
    \cmidrule(lr){1-12}
    & \multirow{2}{*}{Multilingual AV-HuBERT} & A  & \multirow{2}{*}{-} & \multirow{2}{*}{-} & 9.3 & 21.0 & 26.3 & 21.2 & 24.3 & 9.3 & 18.6 \\
    &                                         & AV &                    &                    & 7.6 & 20.5 & 25.2 & 20.0 & 24.0 & 8.1 & 17.6 \\
    \bottomrule
    \end{tabular}
    \label{tab:x_en}
\end{table*}

\section{Corpus Creation}

\muavic~sources data from TED and TEDx talk recordings, where native or non-native speakers (only one speaker most of the time) deliver public speech on stage and cameras capture stage scenes switching among different viewpoints. We collect both audio and video tracks from the recordings, and align them with human transcriptions as well as text translations.

For English talks, we reuse the audio-visual data from LRS3-TED~\cite{afouras2018lrs3} and follow the original data split. We find human translations for these talks from a machine translation corpus, TED2020~\cite{reimers-2020-multilingual-sentence-bert} by matching transcriptions in LRS3-TED and source sentences in TED2020. Matched LRS3-TED examples are then paired with the corresponding target sentences in TED2020 for translation labels. We apply exact text matching for development set and test set examples to ensure the best accuracy. To improve matching recall on the train set, we develop a fuzzy text matching strategy: 
we first segment TED2020 source and target sentences by punctuation if both sides of the sentence pair contain the same amount of segments.
Then we normalize TED2020 and LRS3-TED texts by punctuation removal and lowercasing. Finally, we conduct exact text matching between the two corpora.
For LRS3-TED train set examples without a match from TED2020, we acquire pseudo-translation labels from a machine translation model, M2M-100 \textit{418M}~\cite{fan2021beyond} with default decoding hyper-parameters.

For non-English talks, we reuse the audio-only data, transcriptions and text translations collected by mTEDx~\cite{salesky2021mtedx}. Our data split also follows mTEDx. We acquire video tracks of the original recordings, and align processed video data with the audio one to form audio-visual data, similar to LRS3-TED~\cite{afouras2018lrs3}. Although all the audio data in mTEDx is transcribed, only a subset of it is translated. We acquire pseudo-translation labels from M2M-100 \textit{418M}~\cite{fan2021beyond} for the untranslated train set examples with default decoding hyper-parameters.

In Table~\ref{tab:stats}, we provide duration and speaker statistics for each data split in \muavic.~\muavic~covers 9 languages providing total 1200 hours of transcribed audio-visual data from more than 8000 speakers (number of speakers estimated based on TED/TEDx talk IDs). It also provides text translations into or out of English for 6 of the 9 languages. All the AVSR and AVST test sets are human-labeled, with no less than 0.5 hours of audio-visual speech from at least 7 speakers.


\section{Experiments}
\subsection{Experimental Setup}
For both AVSR and AVST, we use an English AV-HuBERT \textit{large} pre-trained model~\cite{shi22_interspeech}~\footnote{Randomly initialized models perform poorly (see Appendix~\ref{sec:x_en_nopretrain}).}, which is trained on the combination of LRS3-TED~\cite{afouras2018lrs3} and the English portion of VoxCeleb2~\cite{chung2018voxceleb2}. We follow \cite{shi22_interspeech} for fine-tuning hyper-parameters, except that we fine-tune our bilingual models to 30K updates and our multilingual AVSR model to 90K updates. We freeze the pre-trained encoders for the first 4K and 24K updates  for X-En AVST and En-X AVST models, respectively. Also, we randomly augment $25\%$ of the input samples with multiple types of additive noises with a SNR (signal-to-noise ratio) of 0. The noise audio clips in the categories of ``natural'', ``music'' and ``babble'' are sampled from MUSAN dataset~\cite{snyder2015musan}, while the overlapping ``speech'' noise samples are drawn from LRS3-TED. In creating ``speech'' and ``babble'' noise sets, we ensure there are no speaker overlap among different partitions. We remove extremely short utterances (less than 0.2 seconds) and long utterances (more than 20 seconds) for better training stability.
Besides end-to-end AVST models, we also build cascaded systems composed of a AVSR model and a text-based machine translation model (MT). For bilingual MT, we train Transformer \textit{base}~\cite{vaswani2017attention} models using all the transcription-translation pairs in \muavic~train set. For multilingual MT, we leverage M2M-100 \textit{418M}~\cite{fan2021beyond}, which is trained on large-scale open-domain mined bitext data. For speech recognition and X-En speech translation, we also evaluate Whisper \textit{Large V2} ~\cite{radford2022robust} as a SOTA baseline.

For inference, we use the best checkpoint by validation accuracy for AVSR/AVST
and the best checkpoint by validation BLEU for MT. 
We use a beam size of 5 
and default values for the other beam search decoding hyper-parameters. For AVSR, we normalize texts by punctuation removal and lowercasing~\cite{wang-etal-2020-fairseq} before calculating WER (word error rate). For AVST and MT, we use SacreBLEU~\cite{post-2018-call} with default options, where texts are processed by its built-in \textit{13a} tokenizer before BLEU~\cite{papineni2002bleu} calculation. 

\begin{table*}[t]
    \centering
    \fontsize{7.5}{10}\selectfont
    \caption{Results for speech recognition and speech translation in a noisy environment with multilingual babble noise (SNR=0. A: audio, AV: audio+video).}
    \begin{tabular}{lcrrrrrrrrrr}
    \toprule
    \multirow{2}{*}{Model} & \multirow{2}{*}{Mode} & \multicolumn{9}{c}{Source/Target} & \multirow{2}{*}{Avg}\\
    \cmidrule(lr){3-11}
                           &                       & En & Ar & De & El & Es & Fr & It & Pt & Ru \\
    \midrule[\heavyrulewidth]
    \multicolumn{12}{c}{\it Speech Recognition, Test WER $\downarrow$} \\
    Whisper, multilingual, unconstrained data ~\cite{radford2022robust} & A & 202.4 & 197.9 & 244.4 & 113.3 & 116.3 & 172.3 & 172.4 & 223.6 & 126.2 & 174.3 \\
    \cmidrule(lr){1-12}
    \multirow{2}{*}{\centering Monolingual AV-HuBERT}  & A  & 25.9 & 102.3 & 83.5 & 74.6 & 65.1 & 63.6 & 71.0 & 77.3 & 68.1 & 70.2 \\
                                                       & AV & 5.7  & 101.0 & 74.6 & 53.1 & 45.1 & 48.1 & 51.6 & 47.2 & 51.9 & 53.2 \\
    \cmidrule(lr){1-12}
    \multirow{2}{*}{\centering Multilingual AV-HuBERT} & A  & - & 86.1 & 76.0 & 73.7 & 59.7 & 54.7 & 61.0 & 61.9 & 61.9 & 66.9 \\
                                                       & AV & - & 82.2 & 66.9 & 62.2 & 40.7 & 39.0 & 44.3 & 43.1 & 43.1 & 52.7 \\
    
    \midrule[\heavyrulewidth]
    \multicolumn{12}{c}{\it En-X Speech-to-Text Translation, Test BLEU $\uparrow$} \\
    \multirow{2}{*}{\centering Bilingual AV-HuBERT}    & A  & - & - & - & 15.9 & 19.2 & 17.1 & 12.9 & 14.4 & 10.3 & 15.0 \\
                                                       & AV & - & - & - & 22.7 & 24.8 & 23.8 & 20.0 & 20.0 & 13.7 & 20.8 \\
    \midrule[\heavyrulewidth]
    \multicolumn{12}{c}{\it X-En Speech-to-Text Translation, Test BLEU $\uparrow$} \\
    Whisper, multilingual, unconstrained data ~\cite{radford2022robust} & A & - & - & - & 0.1 & 0.4 & 0.7 & 0.1 & 0.1 & 0.2 & 0.3 \\
    \cmidrule(lr){1-12}
    \multirow{2}{*}{\centering Bilingual AV-HuBERT}    & A  & - & - & - & 3.9 & 8.6  & 13.1 & 7.7  & 9.5  & 5.7 & 8.1  \\
                                                       & AV & - & - & - & 4.2 & 12.4 & 16.5 & 12.4 & 15.4 & 6.1 & 11.2 \\
    \cmidrule(lr){1-12}
    \multirow{2}{*}{\centering Multilingual AV-HuBERT} & A  & - & - & - & 2.9 & 8.4  & 12.4 & 8.1  & 8.6  & 0.9 & 6.9  \\
                                                       & AV & - & - & - & 4.2 & 12.8 & 15.0 & 12.5 & 14.8 & 4.6 & 10.7 \\
    
    \bottomrule
    \end{tabular}
    \label{tab:noisy}
\end{table*}

\subsection{Audio-Visual Speech Recognition (AVSR)}
\subsubsection{Clean Setup}
We evaluate AVSR models in both audio-only (``A") and audio-visual (``AV") modes, where the former leverages only audio modality in fine-tuning and inference while the latter leverages both audio and visual modalities. As shown in Table~\ref{tab:en_x}, the English AVSR model~\cite{shi22_interspeech} has a low test WER of 2.5 and 2.3 respectively for audio-only and audio-visual modes. For non-English AVSR, we fine-tune the pre-trained English AV-HuBERT model either separately on each language (8 monolingual models) or jointly on all the 8 non-English languages (a multilingual model), whose test WER can be found in Table~\ref{tab:x_en}. We observe that our monolingual AVSR models in the audio-visual mode outperform a comparable ASR baseline (Transformer, monolingual) by average 52\% WER reduction. They underperform the hybrid LF-MMI monolingual ASR baselines by average 14\% WER increase, which leverage 4-gram language models in decoding while our models do not. Our multilingual AVSR model on average slightly outperforms monolingual AVSR models, with gains observed on some low-resource languages (Ar and De) and degradation observed on the others (El and Ru). We also observe that our multilingual AVSR model falls behind Whisper by average 20\% WER due to the large gap in the amount of training data (0.7K hours compared to 680K hours).

\subsubsection{Noisy Setup}
The first section of Table~\ref{tab:noisy} shows the test WER of our AVSR models in a noisy setup, where we simulate noisy environments by adding multilingual babble noises to clean speech inputs with a SNR (signal-to-noise ratio) of 0 \footnote{See Appendix~\ref{sec:noise_results} for model performance under different noise types and different SNRs}. We observe that Whisper, a SOTA multilingual ASR model, performs catastrophically in this challenging setup, with a high average WER of 174.3 over the 9 languages. In contrast, our monolingual AVSR models in the audio-only mode have an average WER of 70.2 and 66.7 respectively. In the audio-visual mode, the average WER of our models drop significantly by 32\%, suggesting their efficient use of visual information to alleviate the distraction of noisy environments. Our multilingual AVSR model outperforms the monolingual counterparts on every non-English language (except El) in both audio-only and audio-visual modes.

\subsection{Audio-Visual Speech-to-Text Translation (AVST)}
\subsubsection{Clean Setup}
We report test BLEU for En-X AVST and X-En AVST models in Table~\ref{tab:en_x} and Table~\ref{tab:x_en}, respectively. Besides end-to-end AV-HuBERT AVST models, we also set up bilingual and multilingual MT baselines, and build cascaded AVST baselines by pipelining AV-HuBERT AVSR models with MT models. We see that our end-to-end AVST models are on par with the cascaded counterparts in the constrained data setup (``\texttt{A1}+\texttt{A3}'' and ``\texttt{B1}+\texttt{B3}'') for both En-X and X-En directions. Similar to the case in non-English AVSR, our 6-language multilingual X-En AVST model on average perform slightly better than the corresponding bilingual AVST models.

\subsubsection{Noisy Setup}
We evaluate our En-X AVST and X-En AVST models in a noisy setup, whose test BLEU are shown in the second and third section of Table~\ref{tab:noisy}, respectively. We simulate noisy environments in the same approach as that for AVSR models, where multilingual babble noises are added to clean speech inputs with a SNR of 0. We observe that Whisper, a SOTA multilingual X-En speech-to-text translation model, has a catastrophic performance under this setup, with only 0.3 average BLEU over the 6 directions. Our bilingual and multilingual AVST models in the audio-only mode outperform it largely with 7.8 and 6.6 average BLEU improvement, respectively. AVST models in the audio-visual mode consistently outperform those in the audio-only mode, with an improvement of 3.2 and 3.3 on average BLEU for the bilingual and multilingual settings, respectively.

\section{Conclusion}
We introduce a multilingual audio-visual corpus, \muavic, for 9 languages totaling 1200 hours of speech. It is the first open benchmark for audio-visual speech-to-text translation and the largest open benchmark for multilingual audio-visual speech recognition. The corpus is available at \url{https://github.com/facebookresearch/muavic}.

\bibliographystyle{IEEEtran}

\bibliography{mybib}

\clearpage
\appendix

\section{Appendix}


\subsection{Performance of AV-HuBERT without pre-training}
\label{sec:x_en_nopretrain}
\begin{table*}[t]
    \centering
    \fontsize{7.5}{10}\selectfont
    \caption{Performance of randomly initialized AV-HuBERT models for non-English speech recognition and X-En speech-to-text translation
    }
    \begin{tabular}{llcrrrrrrrrrr}
    \toprule
    \multirow{2}{*}{ID} & \multirow{2}{*}{Model} & \multirow{2}{*}{Mode} & \multicolumn{8}{c}{Source} & \multirow{2}{*}{Avg} \\
    \cmidrule(lr){4-11}
     & & & Ar & De & El & Es & Fr & It & Pt & Ru & \\
    \midrule[\heavyrulewidth]
    \multicolumn{12}{c}{\it Speech Recognition, Test WER $\downarrow$} \\
    & \multirow{2}{*}{Monolingual AV-HuBERT, no pre-training} & A  & 99.3 & 61.1 & 35.1 & 16.5 & 24.4 & 19.3 & 23.0 & 33.3 & 39.0 \\
    &                                                         & AV & 98.5 & 52.4 & 26.3 & 15.9 & 23.7 & 18.5 & 19.4 & 30.0 & 35.6 \\
    \midrule[\heavyrulewidth]
    \multicolumn{12}{c}{\it X-En Speech-To-Text Translation, Test BLEU $\uparrow$} \\
    
    & \multirow{2}{*}{Bilingual AV-HuBERT, no pre-training} & A  & \multirow{2}{*}{-} & \multirow{2}{*}{-} & 1.0 & 2.0 & 2.3 & 0.9 & 1.9 & 0.7 & 1.4 \\
    &                                                       & AV &                    &                    & 0.6 & 0.7 & 1.1 & 0.6 & 1.0 & 0.6 & 0.7 \\
    \cmidrule(lr){1-12}
    & \multirow{2}{*}{Multilingual AV-HuBERT, no pre-training} & A  & \multirow{2}{*}{-} & \multirow{2}{*}{-} & 0.6 & 0.9 & 0.7 & 0.7 & 0.8 & 0.8 & 0.7 \\
    &                                                          & AV &                    &                    & 0.7 & 0.7 & 0.7 & 0.6 & 0.9 & 0.8 & 0.7 \\
    \bottomrule
    \end{tabular}
    \label{tab:x_en_nopretrain}
\end{table*}

Table~\ref{tab:x_en_nopretrain} presents the performance of AV-HuBERT models without pre-training on unlabeled audio-visual speech data. Speech recognition and speech-to-text translation performance of randomly initialized models is significantly lower compared to their pre-trained counterparts, as shown in Table~\ref{tab:en_x}. This finding is consistent with the results reported in~\cite{shi22_interspeech} for English-only AVSR.


    \begin{table*}[t]
        \centering
        \fontsize{7}{10}\selectfont
        \caption{Results for speech recognition and speech translation in a noisy environment with multilingual babble noise (SNR=-10, 0 and 10)}
        \begin{tabular}{l@{\hs{0.9}}c@{\hs{0.9}}r@{\hs{0.9}}r@{\hs{0.9}}r@{\hs{0.9}}|@{\hs{0.9}}r@{\hs{0.9}}r@{\hs{0.9}}r@{\hs{0.9}}|@{\hs{0.9}}r@{\hs{0.9}}r@{\hs{0.9}}r@{\hs{0.9}}|@{\hs{0.9}}r@{\hs{0.9}}r@{\hs{0.9}}r@{\hs{0.9}}|@{\hs{0.9}}r@{\hs{0.9}}r@{\hs{0.9}}r@{\hs{0.9}}|@{\hs{0.9}}r@{\hs{0.9}}r@{\hs{0.9}}r@{\hs{0.9}}|@{\hs{0.9}}r@{\hs{0.9}}r@{\hs{0.9}}r}
        \toprule
        \multirow{4}{*}{Model} & \multirow{4}{*}{Mode}  & \multicolumn{3}{c}{El} & \multicolumn{3}{c}{Es} & \multicolumn{3}{c}{Fr} & \multicolumn{3}{c}{It} & \multicolumn{3}{c}{Pt} & \multicolumn{3}{c}{Ru} & \multicolumn{3}{c}{\multirow{3}{*}{Avg by SNR}} \\
        \cmidrule(lr){3-20}
                               &                        & \multicolumn{18}{c}{Multi babble noise, SNR=} & &  \\
                               &                        &    -10 & 0 & 10    &    -10 & 0 & 10    &    -10 & 0 & 10    &    -10 & 0 & 10    &    -10 & 0 & 10    &    -10 & 0 & 10  & -10 & 0 & 10  \\
        \midrule[\heavyrulewidth]
        \multicolumn{23}{c}{\it Speech Recognition, Test WER $\downarrow$} \\
        \multirow{2}{*}{\centering Mono} & A  & 106.1 & 74.6 & 43.1 & 108.5 & 65.1 & 23.6 & 115.0 & 63.6 & 30.3 & 102.4 & 71.0 & 29.8 & 109.1 & 77.3 & 37.3 & 107.6 & 68.1 & 40.4 & 108.1 & 69.9 & 34.1 \\
                                                & AV & 97.4  & 53.1 & 32.2 & 101.9 & 45.1 & 20.7 & 106.6 & 48.1 & 28.3 & 98.7  & 51.6 & 25.0 & 95.9  & 47.2 & 25.4 & 98.3  & 51.9 & 34.1 & 99.8  & 49.5 & 27.6 \\
        \cmidrule(lr){1-23}
        \multirow{2}{*}{\centering Multi} & A  & 97.4 & 73.7 & 47.4 & 94.1 & 59.7 & 35.1 & 92.3 & 54.7 & 30.3 & 94.2 & 61.0 & 27.1 & 92.8 & 61.9 & 28.3 & 96.3 & 64.1 & 43.2 & 94.5 & 62.5 & 35.2 \\
                                                 & AV & 96.0 & 62.2 & 46.1 & 88.6 & 40.7 & 20.8 & 87.0 & 39.0 & 21.9 & 88.0 & 44.3 & 24.5 & 83.9 & 43.1 & 24.8 & 92.9 & 57.9 & 42.9 & 89.4 & 47.9 & 30.2 \\
        
        \midrule[\heavyrulewidth]
        \multicolumn{23}{c}{\it En-X Speech-to-Text Translation, Test BLEU $\uparrow$} \\
        \multirow{2}{*}{\centering Bi} & A  & 0.2  & 15.9 & 22.5 & 0.3  & 19.2 & 26.1 & 0.3  & 17.1 & 23.7 & 0.2  & 12.9 & 19.5 & 0.1 & 14.4 & 19.6 & 0.3  & 10.3 & 14.1 & 0.2  & 15.0 & 20.9 \\
                                              & AV & 12.8 & 22.7 & 23.0 & 15.3 & 24.8 & 26.1 & 12.1 & 23.8 & 24.7 & 10.4 & 20.0 & 20.4 & 11.5 & 20.0 & 20.6 & 7.8 & 13.7 & 14.6 & 11.6 & 20.8 & 21.6 \\
        
        \midrule[\heavyrulewidth]
        \multicolumn{23}{c}{\it X-En Speech-to-Text Translation, Test BLEU $\uparrow$} \\
        \multirow{2}{*}{\centering Bi}    & A  & 0.6 & 3.9 & 6.4 & 0.3 & 8.6  & 17.9 & 0.6 & 13.1 & 24.5 & 0.6 & 7.7  & 17.2 & 0.6 & 9.5  & 20.3 & 0.4 & 5.7 & 8.7 & 0.5 & 8.1  & 15.8 \\
                                                 & AV & 1.6 & 4.2 & 6.5 & 2.2 & 12.4 & 18.4 & 3.2 & 16.5 & 23.5 & 3.2 & 12.4 & 17.6 & 4.4 & 15.4 & 22.0 & 1.6 & 6.1 & 8.6 & 2.7 & 11.2 & 16.1 \\
        \cmidrule(lr){1-23}
        \multirow{2}{*}{\centering Multi} & A  & 0.4 & 2.9 & 7.6 & 0.5 & 8.4  & 18.3 & 0.4 & 12.4 & 23.8 & 0.4 & 8.1  & 17.7 & 0.5 & 8.6  & 19.9 & 0.4 & 3.9 & 7.8 & 0.4 & 7.4  & 15.9 \\
                                                 & AV & 0.6 & 4.2 & 6.5 & 2.1 & 12.8 & 18.7 & 2.4 & 15.0 & 23.4 & 2.3 & 12.5 & 18.4 & 3.4 & 14.8 & 21.8 & 0.7 & 4.6 & 6.9 & 1.9 & 10.7 & 16.0 \\
        \bottomrule
        \end{tabular}
        \label{tab:noisy_multi_babble}
    \end{table*}

    \begin{table*}[t]
        \centering
        \fontsize{7}{10}\selectfont
        \caption{Results for speech recognition and speech translation in a noisy environment with speech noise (SNR=-10, 0 and 10)}
        \begin{tabular}{l@{\hs{0.9}}c@{\hs{0.9}}r@{\hs{0.9}}r@{\hs{0.9}}r@{\hs{0.9}}|@{\hs{0.9}}r@{\hs{0.9}}r@{\hs{0.9}}r@{\hs{0.9}}|@{\hs{0.9}}r@{\hs{0.9}}r@{\hs{0.9}}r@{\hs{0.9}}|@{\hs{0.9}}r@{\hs{0.9}}r@{\hs{0.9}}r@{\hs{0.9}}|@{\hs{0.9}}r@{\hs{0.9}}r@{\hs{0.9}}r@{\hs{0.9}}|@{\hs{0.9}}r@{\hs{0.9}}r@{\hs{0.9}}r@{\hs{0.9}}|@{\hs{0.9}}r@{\hs{0.9}}r@{\hs{0.9}}r}
        \toprule
        \multirow{4}{*}{Model} & \multirow{4}{*}{Mode}  & \multicolumn{3}{c}{El} & \multicolumn{3}{c}{Es} & \multicolumn{3}{c}{Fr} & \multicolumn{3}{c}{It} & \multicolumn{3}{c}{Pt} & \multicolumn{3}{c}{Ru} & \multicolumn{3}{c}{\multirow{3}{*}{Avg by SNR}} \\
        \cmidrule(lr){3-20}
                               &                        & \multicolumn{18}{c}{Speech noise, SNR=} & &  \\
                               &                        &    -10 & 0 & 10    &    -10 & 0 & 10    &    -10 & 0 & 10    &    -10 & 0 & 10    &    -10 & 0 & 10    &    -10 & 0 & 10  & -10 & 0 & 10  \\
        \midrule[\heavyrulewidth]
        \multicolumn{23}{c}{\it Speech Recognition, Test WER $\downarrow$} \\
        \multirow{2}{*}{\centering Mono} & A  & 78.6 & 51.7 & 39.9 & 60.5 & 29.5 & 20.3 & 72.2 & 38.2 & 27.8 & 69.8 & 38.1 & 24.8 & 76.3 & 47.1 & 32.3 & 74.8 & 48.8 & 37.9 & 72.0 & 42.2 & 30.5 \\
                                                & AV & 58.1 & 36.8 & 30.1 & 47.1 & 25.7 & 18.9 & 54.8 & 32.5 & 26.2 & 53.8 & 30.5 & 22.7 & 49.0 & 29.5 & 23.5 & 58.4 & 38.4 & 32.6 & 53.5 & 32.2 & 25.7 \\
        \cmidrule(lr){1-23}
        \multirow{2}{*}{\centering Multi} & A  & 72.4 & 53.0 & 44.2 & 60.0 & 39.9 & 34.3 & 56.2 & 37.5 & 29.6 & 60.9 & 34.3 & 23.9 & 59.1 & 33.9 & 25.2 & 67.6 & 50.0 & 41.3 & 62.7 & 41.4 & 33.1 \\
                                                 & AV & 63.8 & 50.5 & 44.0 & 42.5 & 24.3 & 18.7 & 42.8 & 25.4 & 20.9 & 45.8 & 28.6 & 22.6 & 44.4 & 28.5 & 23.2 & 60.5 & 46.6 & 41.6 & 50.0 & 34.0 & 28.5 \\
        
        \midrule[\heavyrulewidth]
        \multicolumn{23}{c}{\it En-X Speech-to-Text Translation, Test BLEU $\uparrow$} \\
        \multirow{2}{*}{\centering Bi} & A  & 3.9  & 14.2 & 20.9 & 5.9  & 17.9 & 25.1 & 5.2  & 15.3 & 22.3 & 3.8  & 12.2 & 18.6 & 3.9  & 12.7 & 18.2 & 3.3  & 9.0  & 13.6 & 4.3  & 13.6 & 19.8 \\
                                              & AV & 20.6 & 23.2 & 22.9 & 23.5 & 25.8 & 26.6 & 21.8 & 24.4 & 24.9 & 18.1 & 20.3 & 20.3 & 18.8 & 20.4 & 20.5 & 12.4 & 14.4 & 14.4 & 19.2 & 21.4 & 21.6 \\
        
        \midrule[\heavyrulewidth]
        \multicolumn{23}{c}{\it X-En Speech-to-Text Translation, Test BLEU $\uparrow$} \\
        \multirow{2}{*}{\centering Bi}    & A  & 3.4 & 6.0 & 6.8 & 8.4  & 15.8 & 19.1 & 11.1 & 21.3 & 25.0 & 7.0  & 15.1 & 17.6 & 8.8  & 17.6 & 21.1 & 4.3 & 7.5 & 8.7 & 7.2 & 13.9 & 16.4 \\
                                                 & AV & 4.0 & 5.7 & 6.7 & 11.1 & 16.8 & 19.2 & 14.0 & 21.2 & 24.0 & 11.0 & 16.8 & 18.5 & 13.7 & 20.2 & 22.7 & 5.4 & 8.0 & 8.6 & 9.8 & 14.8 & 16.6 \\
        \cmidrule(lr){1-23}
        \multirow{2}{*}{\centering Multi} & A  & 2.8 & 6.7 & 8.7 & 8.3  & 16.4 & 19.3 & 10.7 & 20.9 & 24.7 & 7.6  & 15.9 & 19.1 & 8.5  & 17.7 & 21.7 & 3.4 & 6.6 & 8.2 & 6.9 & 14.0 & 16.9 \\
                                                 & AV & 4.0 & 6.2 & 6.8 & 12.1 & 17.9 & 19.5 & 14.0 & 21.3 & 23.9 & 11.4 & 16.9 & 19.0 & 14.1 & 20.1 & 22.5 & 3.7 & 6.0 & 7.5 & 9.9 & 14.7 & 16.5 \\
        \bottomrule
        \end{tabular}
        \label{tab:noisy_speech}
    \end{table*}

    \begin{table*}[t]
        \centering
        \fontsize{7.5}{10}\selectfont
        \caption{Results for speech recognition and speech translation in a noisy environment with music noise (SNR=-10, 0 and 10)}
        \begin{tabular}{l@{\hs{0.9}}c@{\hs{0.9}}r@{\hs{0.9}}r@{\hs{0.9}}r@{\hs{0.9}}|@{\hs{0.9}}r@{\hs{0.9}}r@{\hs{0.9}}r@{\hs{0.9}}|@{\hs{0.9}}r@{\hs{0.9}}r@{\hs{0.9}}r@{\hs{0.9}}|@{\hs{0.9}}r@{\hs{0.9}}r@{\hs{0.9}}r@{\hs{0.9}}|@{\hs{0.9}}r@{\hs{0.9}}r@{\hs{0.9}}r@{\hs{0.9}}|@{\hs{0.9}}r@{\hs{0.9}}r@{\hs{0.9}}r@{\hs{0.9}}|@{\hs{0.9}}r@{\hs{0.9}}r@{\hs{0.9}}r}
        \toprule
        \multirow{4}{*}{Model} & \multirow{4}{*}{Mode}  & \multicolumn{3}{c}{El} & \multicolumn{3}{c}{Es} & \multicolumn{3}{c}{Fr} & \multicolumn{3}{c}{It} & \multicolumn{3}{c}{Pt} & \multicolumn{3}{c}{Ru} & \multicolumn{3}{c}{\multirow{3}{*}{Avg by SNR}} \\
        \cmidrule(lr){3-20}
                               &                        & \multicolumn{18}{c}{Music noise, SNR=} & &  \\
                               &                        &    -10 & 0 & 10    &    -10 & 0 & 10    &    -10 & 0 & 10    &    -10 & 0 & 10    &    -10 & 0 & 10    &    -10 & 0 & 10  & -10 & 0 & 10  \\
        \midrule[\heavyrulewidth]
        \multicolumn{23}{c}{\it Speech Recognition, Test WER $\downarrow$} \\
        \multirow{2}{*}{\centering Bi} & A  & 76.2 & 49.3 & 39.1 & 68.8 & 31.1 & 19.2 & 70.9 & 37.3 & 26.9 & 73.3 & 37.9 & 24.0 & 77.8 & 43.5 & 28.9 & 74.8 & 47.1 & 37.5 & 73.6 & 41.0 & 29.2 \\
                                              & AV & 64.7 & 37.1 & 29.1 & 53.1 & 25.7 & 17.8 & 55.4 & 32.7 & 25.7 & 58.0 & 31.1 & 21.7 & 54.2 & 29.9 & 22.2 & 60.9 & 39.9 & 32.1 & 57.7 & 32.7 & 24.8 \\
        \cmidrule(lr){1-23}
        \multirow{2}{*}{\centering Multi} & A  & 74.3 & 52.4 & 43.8 & 63.5 & 40.1 & 33.3 & 58.9 & 36.5 & 29.7 & 63.6 & 34.1 & 23.0 & 62.8 & 35.0 & 23.4 & 69.2 & 48.6 & 40.4 & 65.4 & 41.1 & 32.3 \\
                                                 & AV & 66.7 & 50.1 & 43.5 & 45.1 & 25.3 & 18.2 & 45.6 & 26.1 & 20.5 & 49.4 & 29.1 & 22.2 & 47.8 & 29.3 & 22.5 & 63.6 & 46.8 & 40.2 & 53.1 & 34.4 & 27.9 \\
        
        \midrule[\heavyrulewidth]
        \multicolumn{23}{c}{\it En-X Speech-to-Text Translation, Test BLEU $\uparrow$} \\
        \multirow{2}{*}{\centering Bi} & A  & 11.2 & 21.4 & 22.3 & 14.8 & 24.8 & 26.6 & 13.0 & 22.8 & 24.5 & 10.3 & 18.0 & 20.0 & 10.2 & 19.0 & 19.8 & 8.3  & 13.7 & 14.7 & 11.3 & 19.9 & 21.3 \\
                                              & AV & 20.6 & 22.9 & 23.1 & 23.7 & 25.9 & 26.2 & 22.0 & 24.4 & 24.8 & 18.6 & 20.6 & 20.4 & 18.7 & 20.5 & 20.6 & 12.6 & 14.5 & 14.5 & 19.4 & 21.4 & 21.6 \\
        
        \midrule[\heavyrulewidth]
        \multicolumn{23}{c}{\it X-En Speech-to-Text Translation, Test BLEU $\uparrow$} \\
        \multirow{2}{*}{\centering Bi}    & A  & 3.3 & 6.0 & 6.7 & 7.6  & 15.9 & 19.3 & 11.2 & 22.1 & 25.6 & 7.6  & 15.3 & 18.6 & 8.8  & 18.1 & 22.4 & 4.4 & 7.5 & 8.7 & 7.1 & 14.1 & 16.9 \\
                                                 & AV & 3.5 & 5.8 & 6.7 & 10.2 & 17.0 & 19.2 & 13.4 & 21.5 & 24.3 & 11.0 & 16.2 & 18.7 & 14.0 & 20.2 & 22.7 & 4.9 & 7.8 & 8.8 & 9.5 & 14.7 & 16.7 \\
        \cmidrule(lr){1-23}
        \multirow{2}{*}{\centering Multi} & A  & 3.0 & 6.4 & 8.3 & 7.5  & 16.3 & 19.6 & 10.4 & 20.9 & 25.2 & 7.8  & 15.9 & 19.5 & 7.8  & 17.9 & 22.3 & 3.1 & 6.7 & 8.4 & 6.6 & 14.0 & 17.2 \\
                                                 & AV & 3.9 & 6.4 & 7.1 & 10.7 & 17.3 & 19.5 & 13.3 & 21.0 & 24.4 & 10.7 & 16.9 & 19.2 & 12.8 & 20.2 & 23.2 & 3.3 & 6.6 & 7.4 & 9.1 & 14.7 & 16.8 \\
        \bottomrule
        \end{tabular}
        \label{tab:noisy_music}
    \end{table*}
\subsection{Results for AVSR and AVST models in noisy environments with other noise types and SNRs}
\label{sec:noise_results}

In Table~\ref{tab:noisy_multi_babble}, \ref{tab:noisy_speech} and \ref{tab:noisy_music}, we provide more comprehensive results for our AVSR and AVST models in noisy environments with 3 noise types (multilingual babble noise, speech noise and music noise) and 3 different SNRs (-10, 0 and 10). We observe that AVSR and AVSR models in the audio-visual mode consistently outperforms those in the audio-only mode across all the noise types and SNRs.

\end{document}